\documentclass[twoside]{article}
\usepackage{aistats2016}

\usepackage{mathtools}
\usepackage{dsfont}
\usepackage{amssymb}
\usepackage{amsmath}
\usepackage{graphicx}

\begin{document}



\twocolumn[

\aistatstitle{Exploring the Entire Regularization Path for \\the Asymmetric Cost Linear Support Vector Machine}

%----> In fact, L1-LSVM?

\aistatsauthor{ Daniel Wesierski }

\aistatsaddress{ Gdansk University of Technology \\ Systems Research Institute, Polish Academy of Sciences } ]

\begin{abstract}
We propose an algorithm for exploring the entire regularization path of asymmetric-cost linear support vector machines. Empirical evidence suggests the predictive power of support vector machines depends on the regularization parameters of the training algorithms. The algorithms exploring the entire regularization paths have been proposed for single-cost support vector machines thereby providing the complete knowledge on the behavior of the  trained model over the hyperparameter space. Considering the problem in two-dimensional hyperparameter space though enables our algorithm to maintain greater flexibility in dealing with special cases and sheds light on problems encountered by algorithms building the paths in one-dimensional spaces. We demonstrate two-dimensional regularization paths for linear support vector machines that we train on synthetic and real data.
\end{abstract}

\section{Introduction}

Support Vector Machines (Boser et al., 1992) belong to core machine learning techniques for binary classification. Given a large number of training samples characterized by a large number of features, a linear SVM is often the \textit{go-to} approach in many applications. A handy collection of software packages, e.g., \texttt{LIBLINEAR} (Fan et al., 2008), \texttt{Pegasos} (Shalev-Shwartz et al., 2011), \texttt{SVM$^{\texttt{perf}}$} (Joachims, 2006), \texttt{Scikit-learn} (Pedregosa et al., 2011) provide practitioners with efficient algorithms for fitting linear models to datasets. Finding optimal hyperparameters of the algorithms for model selection is crucial though for good performance at test-time. 

A vanilla cross-validated grid-search is the most common approach to choosing satisfactory hyperparameters. However, grid search scales exponentially with the number of hyperparameters while choosing the right sampling scheme over the hyperparameter space impacts model performance (Bergstra \& Bengio, 2012). Linear SVMs typically require setting a single $C$ hyperparameter that equally regularizes the training \texttt{loss} of misclassified data. (Klatzer \& Pock, 2015) propose bi-level optimization for searching several hyperparameters of linear and kernel SVMs and (Chu et al., 2015) use warm-start techniques to efficiently fit an SVM to large datasets but both approaches explore the hyperparameter regularization space partially. 

The algorithm proposed in (Hastie et al., 2004) builds the entire regularization path for linear and kernel SVMs that use single, symmetric cost for misclassifying negative and positive data\footnote{Solutions path algorithms relate to parametric programming techniques. With independent revival in machine learning, these techniques have traditionally been  applied in optimization and control theory (Gartner et al., 2012)}. The stability of the algorithm was improved in (Ong et al., 2010) by augmenting the search space of feasible event updates from one- to multi-dimensional hyperparameter space. In this paper, we also show that a one-dimensional path following method can diverge to unoptimal solution wrt KKT conditions.

Many problems often require setting multiple hyperparameters (Karasuyama et al., 2012). They arise especially when dealing with imbalanced datasets (Japkowicz \& Stephen, 2002) and require training an SVM with two cost hyperparameters assymetrically attributed to positive and negative examples. (Bach et al., 2006) builds a pencil of one-dimensional regularization paths for the assymetric-cost SVMs. On the other hand, (Karasuyama et al., 2012) build a one-dimensional  regularization path but in a multidimensional hyperspace.

In contrast to algorithms building one-dimensional paths in higher-dimensional hyperparameter spaces, we describe a solution path algorithm that explores the entire regularization path for an assymetric-cost linear SVMs. Hence, our path is a two-dimensional path in the two-dimensional hyperparameter space. 

Our main contributions include: 
\begin{itemize}
\item development of the entire regularization path for assymetric-cost linear support vector machine (AC-LSVM)
\item algorithm initialization at arbitrary location in the $(C^+,C^-)$ hyperparameter space 
\item computationally and memory efficient algorithm amenable to local parallelization.
\end{itemize}

\section{Problem formulation}

Our binary classification task requires a \textit{fixed} input set of $N$ training examples $\forall_{i\in \mathbb{N}_i} \mathbf{x}_i$, where $\mathbf{x}_i \in \mathbb{R}^{d\times 1}$, $\mathcal{N}_i = \lbrace i: i \in \mathbb{N}^+ \wedge i \in \langle 1,N \rangle \rbrace$, $d \in \mathbb{N}^+$, to be annotated with corresponding binary labels $y_i \in \lbrace -1, +1 \rbrace$ denoting either class. Then, the objective is to learn a decision function $g(\mathbf{x}_t)$ that will allow its associated classifier $y_t\leftarrow\texttt{sign}\left[ g(\mathbf{x}_t)\right] $ to predict the label $y_t$ for new sample $\mathbf{x}_t$ at test-time. 

AC-LSVM learns the array of parameters $\boldsymbol{\beta} \in \mathbb{R}^{d\times 1}$ of the decision function $g(\mathbf{x}_i) = \boldsymbol{\beta}^T \mathbf{x}_i$ by solving the following, primal quadratic program (QP):
\begin{align}
\label{eq:linsvm}
&\underset{\boldsymbol{\beta}, \mathbf{\xi}}{\text{argmin}}~~\dfrac{1}{2} \Vert \boldsymbol{\beta}\Vert_2^2 + C^+\sum_{i=1}^{N^+}\xi_i + C^-\sum_{i=1+N^{+}}^{N}\xi_i\\
&~~~~\text{s.t.} ~~ \forall_{i} ~ \boldsymbol{\beta}^T \mathbf{x}_i \geq 1 - \xi_i, ~~\xi_i \geq 0
\end{align} 
where we include the scalar valued bias term $b$ in $\boldsymbol{\beta}^T$ and augment data points $\mathbf{x}_i^T$ by some constant $B$: 
\begin{equation}
\boldsymbol{\beta}^T \leftarrow \left[ \widehat{\boldsymbol{\beta}}^T, b\right], ~~~~  \mathbf{x}_i^T \leftarrow y_i \left[ \widehat{\mathbf{x}}_i^T, B\right]
\label{eq:bias_augm}
\end{equation}
where $B$ is defined by a user (Hsieh, 2008). The above formulation should learn $\boldsymbol{\beta}$ to assign scores higher than margin $1$ to positive examples $\left\lbrace x_i,y_i=+1\right\rbrace $ and lower than margin $-1$ to negative examples $\left\lbrace x_i,y_i=-1\right\rbrace $. As data may be inseparable in $\mathbb{R}^{d\times 1}$, the objective function (1) penalizes violations of these constraints (2) with slack variables $\xi_i \geq 0$, asymmetrically weighted by constants $C^+$ and $C^-$. 

\paragraph{Active sets} Solving the primal QP \eqref{eq:linsvm} is often approached with the help of Lagrange multipliers $\boldsymbol{\alpha} \in \mathbb{R}^N$, where
$\boldsymbol{\alpha} = \left[ \alpha_1,\ldots,\alpha_N \right]^T$, which are associated with $N$ constraints in \eqref{eq:linsvm}. Let $\mathbf{X}_{d \times N} = \left[ \mathbf{x}_1,\ldots,\mathbf{x}_N \right]$ and $\mathbf{1} = \left[ 1, \ldots, 1\right]^T $. Then, the dual problem takes the familiar form:
\begin{align}
&\underset{\boldsymbol{\alpha}}{\text{argmin}}~~\dfrac{1}{2} \boldsymbol{\alpha}^T\mathbf{X}^T\mathbf{X}\boldsymbol{\alpha} - \boldsymbol{\alpha}^T \mathbf{1}\\
&\text{s.t.} ~~ \forall_{i\in \langle 1,N^+ \rangle} ~~~~~ 0 \leq \alpha_i \leq C^+ \\
&~~~~~~\forall_{i\in \langle 1+N^+,N \rangle} ~ 0 \leq \alpha_i \leq C^-
\end{align} 

The immediate consequence of applying the Lagrange multipliers is the expression for the LSVM parameters $\boldsymbol{\beta}=\mathbf{X}\boldsymbol{\alpha}$ yielding the decision function $g(\mathbf{x}_i) = \boldsymbol{\beta}^T \mathbf{x}_i = \mathbf{x}_i^T\mathbf{X}\boldsymbol{\alpha}$. 

The optimal solution $\boldsymbol{\alpha}$ of the dual problem is dictated by satisfying the usual Karush-Kuhn-Tucker (KKT) conditions. Notably, the KKT conditions can be algebraically rearranged giving rise to the following \textit{active sets}:
\begin{equation}
\mathcal{M}^{+/-} = \lbrace i: g(\mathbf{x}_i) = \mathbf{x}_i^T\mathbf{X}\boldsymbol{\alpha} = 1, 0 \leq \alpha_i \leq C^{+/-} \rbrace 
\label{eq:Mset}
\end{equation}
\begin{equation}
\mathcal{I}^{+/-} = \lbrace i: g(\mathbf{x}_i) = \mathbf{x}_i^T\mathbf{X}\boldsymbol{\alpha} < 1, \alpha_i = C^{+/-} \rbrace 
\label{eq:Lset}
\end{equation}
\begin{equation}
\mathcal{O} = \lbrace i: g(\mathbf{x}_i) = \mathbf{x}_i^T\mathbf{X}\boldsymbol{\alpha} > 1, \alpha_i = 0 \rbrace 
\label{eq:Rset}
\end{equation}
Firstly, the $\mathcal{M}\mathcal{I}\mathcal{O}$ sets \eqref{eq:Mset}$-$\eqref{eq:Rset} cluster data points $\mathbf{x}_i$ to the margin $\mathcal{M}$, to the left $\mathcal{I}$, and to the right $\mathcal{O}$ of the margin along with their associated scores $g(\mathbf{x}_i)$. Secondly, the sets indicate the range within the $(C^+,C^-)$ space for Lagrange multipliers $\alpha_i$ over which $\boldsymbol{\alpha}$ is allowed to vary thereby giving rise to a convex polytope in that space.

\paragraph{Convex polytope}

A unique region in $(C^+,C^-)$ satisfying a particular configuration of the $\mathcal{M}\mathcal{I}\mathcal{O}$ set is bounded by a convex polytope. The first task in path exploration is thus to obtain the boundaries of the convex polytope. Following (Hastie, 2004), we obtain\footnote{A similar derivation appeared in (Bach et al., 2006).} linear inequality constraints from \eqref{eq:Mset}$-$\eqref{eq:Rset}: 
\begin{equation}
\mathbf{h}_{\alpha_0^{+/-}}^T=\left[ -\mathbf{X}_{\mathcal{M}}^*\mathbf{X}_{\mathcal{I}_{+}} \mathbf{1}~~
-\mathbf{X}_{\mathcal{M}}^*\mathbf{X}_{\mathcal{I}_{-}} \mathbf{1}~~
(\mathbf{X}_{\mathcal{M}}^T\mathbf{X}_{\mathcal{M}})^{-1}\mathbf{1}\right] 
\label{eq:h_alpha0}
\end{equation}
\begin{equation}
\mathbf{h}_{\alpha_C^+}^T=  \mathbf{h}_{\alpha_0^+}^T + \left[\mathbf{1} ~~ \mathbf{0} ~~ \mathbf{0}\right]^T
\label{eq:h_alphaC1}
\end{equation}
\begin{equation}
\mathbf{h}_{\alpha_C^-}^T= \mathbf{h}_{\alpha_0^-} + \left[\mathbf{0} ~~ \mathbf{1} ~~ \mathbf{0}\right]^T
\label{eq:h_alphaC2}
\end{equation}
\begin{equation}
\mathbf{h}_{\mathcal{I}}^T=\left[ -\mathbf{x}_{\mathcal{I}}^T \mathbf{P}_{\mathcal{M}}^{\perp}~\mathbf{x}_{\mathcal{I}_+} \mathbf{1}~~
-\mathbf{x}_{\mathcal{L}}^T \mathbf{P}_{\mathcal{M}}^{\perp}~\mathbf{x}_{\mathcal{I}_-} \mathbf{1}~~
\mathbf{1}-\mathbf{x}_{\mathcal{I}}^T~\mathbf{X}_{\mathcal{M}}^{\star T} \mathbf{1}\right] 
\label{eq:h_L}
\end{equation}
\begin{equation}
\mathbf{h}_{\mathcal{O}}^T=\left[ \mathbf{x}_{\mathcal{O}}^T \mathbf{P}_{\mathcal{M}}^{\perp}~\mathbf{x}_{\mathcal{I}_{+}} \mathbf{1}~~
\mathbf{x}_{\mathcal{O}}^T \mathbf{P}_{\mathcal{M}}^{\perp}~\mathbf{x}_{\mathcal{I}_{-}} \mathbf{1}~~
-(\mathbf{1}-\mathbf{x}_{\mathcal{O}}^T~\mathbf{X}_{\mathcal{M}}^{\star T} \mathbf{1})\right] 
\label{eq:h_R}
\end{equation}
where, $\mathbf{P}_{\mathbf{X}^{\perp}_{\mathcal{M}}}=\mathbf{I} - \mathbf{X}_{\mathcal{M}} \mathbf{X}_{\mathcal{M}}^*$ is the orthogonal projector onto the orthogonal complement of the subspace spanned by $\mathbf{X}_{\mathcal{M}}$ and $\mathbf{X}_{\mathcal{M}}^*=(\mathbf{X}_{\mathcal{M}}^T\mathbf{X}_{\mathcal{M}})^{-1} \mathbf{X}_{\mathcal{M}}^T$ is the Moore-Penrose pseudoinverse if $\mathbf{X}_{\mathcal{M}}$ has full column rank.

Specifically, let $\mathbf{H}$ be a matrix composed of constraints \eqref{eq:h_alpha0}$-$\eqref{eq:h_R}. 
\begin{equation}
\mathbf{H}=\left[ \mathbf{h}_{\alpha_0^+} ~~ \mathbf{h}_{\alpha_0^-} ~~ \mathbf{h}_{\alpha_C^+} ~~ \mathbf{h}_{\alpha_C^-} ~~ \mathbf{h}_{\mathcal{L}} ~~ \mathbf{h}_{\mathcal{R}} \right]
\label{eq:H}
\end{equation}
Then, the boundaries of the convex polytope in the $(C^+,C^-)$ space are indicated by a subset of active constraints in $\mathbf{c}^T \mathbf{H} \geq \mathbf{0}^T$, which evaluate to $0$ for some $\mathbf{c}^T = \left[C^+ ~~ C^- ~~ 1\right]$. The boundaries can be determined in linear time $\mathcal{O}(N)$ with efficient convex hull (\texttt{CH}) routines (Avis et al., 1997).

Now, in order grow the entire regularization path in $(C^+,C^-)$, the $\mathcal{M}\mathcal{I}\mathcal{O}$ sets have to updated at $l$-th step such that the KKT conditions will hold, thereby determining an $(l+1)$-th convex polytope. The polytope constraints $\mathbf{h}_{\alpha_0^{+/-}}$ for which $\alpha_i = 0$ indicate that a point $\mathbf{x}_i$ has to go from $\mathcal{M}$ to $\mathcal{O}$ in order to satisfy the KKT conditions. Likewise, $\mathbf{h}_{\alpha_C^{+/-}}$ for which $\alpha_i = C^{+/-}$ indicate updating the point from $\mathcal{M}$ to $\mathcal{I}$, $\mathbf{h}_{\mathcal{R}}$ indicate point transition from $\mathcal{O}$ to $\mathcal{M}$, and $\mathbf{h}_{\mathcal{I}}$ indicate point transition from $\mathcal{I}$ to $\mathcal{M}$. These set transitions are usually called events, while the activated constraints are called breakpoints. 

Therefore, at breakpoint, we determine the event for $i$-th point by a function that updates the $\mathcal{M}_l\mathcal{I}_l\mathcal{O}_l$ set from $l$ to $l+1$ as: 
\begin{equation}
u(i,t,l) = \begin{cases} 
\mathcal{M}_{l+1}= \mathcal{M}_{l} \setminus i \wedge \mathcal{O}_{l+1}= \mathcal{O}_{l} \cup i,& t = 0 \\
\mathcal{O}_{l+1}= \mathcal{O}_{l} \setminus i \wedge \mathcal{M}_{l+1}= \mathcal{M}_{l} \cup i,& t = 0 \\
\mathcal{M}_{l+1}= \mathcal{M}_{l} \setminus i \wedge \mathcal{I}_{l+1}= \mathcal{I}_{l} \cup i,& t = 1 \\
\mathcal{I}_{l+1}= \mathcal{I}_{l} \setminus i \wedge \mathcal{M}_{l+1}= \mathcal{M}_{l} \cup i,& t = 1 \\
\end{cases}
\label{eq:update}
\end{equation}
where the direction of the transition depends on the current $\mathcal{M}_l\mathcal{I}_l\mathcal{O}_l$ set configuration.

Following (Hastie et al., 2004), our algorithm requires $\mathcal{M}_l\mathcal{I}_l\mathcal{O}_l$ set to proceed to the next set by \eqref{eq:update}. However, unlike (Hastie et al., 2004), the constraints \eqref{eq:h_alpha0}$-$\eqref{eq:h_R} are independent of the previous computations of $\alpha_i$ and $\boldsymbol{\beta}$. This has several implications. Firstly, our algorithm does not accumulate potential numerical errors in these parameters. Secondly, the algorithm can be initialized from an arbitrary location in the $(C^+,C^-)$ space.

\section{Proposed method}
\label{sec:algo}

The evolution of $\boldsymbol{\alpha}$ is continuous and piecewise linear in the $(C^+,C^-)$ space (Bach et al., 2006). An immediate consequence is that the active constraints have to flip during the set update \eqref{eq:update}.

\paragraph{Flipping constraints} Suppose we have a single event that $i \in \mathcal{M}_l^+$ goes to $\mathcal{I}_{l+1}^+$ thereby forcing the set update rule $u(i,t=1,l)$. Then,  the $i$-th constraint in \eqref{eq:h_alpha0} can be rearranged using the \textit{matrix inversion lemma} wrt $\mathbf{x}_i$ as:
\begin{equation}
\begin{bmatrix}
\mathbf{x}_i^T \mathbf{P}_{\mathbf{X}^{\perp}_{\mathcal{M}_l \setminus i}} \mathbf{X}_{\mathcal{I}^{+}_{l}} \mathbf{1} + \mathbf{x}_i^T \mathbf{P}_{\mathbf{X}^{\perp}_{\mathcal{M}_l \setminus i}} \mathbf{x}_i\\
\mathbf{x}_i^T \mathbf{P}_{\mathbf{X}^{\perp}_{\mathcal{M}_l \setminus i}} \mathbf{X}_{\mathcal{I}^{-}_{l}} \mathbf{1} \\           
-1+\mathbf{x}_i^T~\mathbf{X}_{\mathcal{M}_l \setminus i}^{\star T} \mathbf{1}
\end{bmatrix}
\label{eq:h_alpha0_inv}
\end{equation}
where $\mathbf{x}_i^T \mathbf{P}_{\mathbf{X}^{\perp}_{\mathcal{M}_l \setminus i}} \mathbf{x}_i \geq 0 $. This constraint is equal to its corresponding, sign-flipped counterpart in \eqref{eq:h_L} at $l+1$ as: 
\begin{equation}
\begin{bmatrix}
-\mathbf{x}_i^T \mathbf{P}_{\mathbf{X}^{\perp}_{\mathcal{M}_{l+1}}} (\mathbf{X}_{\mathcal{I}^{+}_{l}} \mathbf{1} + \mathbf{x}_i)\\
-\mathbf{x}_i^T \mathbf{P}_{\mathbf{X}^{\perp}_{\mathcal{M}_{l+1}}} \mathbf{X}_{\mathcal{I}^{-}_{l+1}} \mathbf{1} \\           
1-\mathbf{x}_i^T~\mathbf{X}_{\mathcal{M}_{l+1}}^{\star T} \mathbf{1}
\end{bmatrix}
\label{eq:h_L_inv}
\end{equation}
with $\mathcal{I}^{-}_{l} = \mathcal{I}^{-}_{l+1}$. The same argument holds for update type $t=0$. 

Furthermore, (Hastie et al., 2004) express the evolution of $\boldsymbol{\alpha}$ with a single cost parameter $C$. This case is equivalent to $C^-=C^+$ that yields the identity line in the $(C^+,C^-)$ space. (Ong et al., 2010) observe that one-dimensional path exploration over a line can lead to incorrect results and resort to searching for alternative set updates over a higher-dimensional hyperparameter space. Notably, when two points hit the margin at the same time at $l$, the matrix updated by both points $\mathbf{X}_{\mathcal{M}_{l+1}}^T\mathbf{X}_{\mathcal{M}_{l+1}}$ not necessarily needs to become singular. However, the $\mathcal{M}_l\mathcal{I}_l\mathcal{O}_l$ sets can be incorrectly updated. We formalize this by introducing the notion of \textit{joint events} that may co-occur at some point on the line. In our setting of the $2$D path exploration, this is always the case when a vertex of a polytope coincides with a line in the $(C^+,C^-)$ space.

\paragraph{Joint events} At the vertices of the convex polytope at least two events occur concurrently. In this case, the $\mathcal{M}\mathcal{I}\mathcal{O}$ set can be updated twice from $l$ to $l+1$. Hence, this vertex calls $3$ different updates of the $\mathcal{M}_l\mathcal{I}_l\mathcal{O}_l$ set, \textit{i.e.} two single updates for both edges and a joint update. 

Note that the piecewise continuous $2$D path of $\boldsymbol{\alpha}$ also implies piecewise continuous $1$D path of the events. Moreover, as each vertex is surrounded by $4$ different $\mathcal{M}\mathcal{I}\mathcal{O}$ sets, two events at the vertex have to satisfy the following \textit{vertex loop property}:
\begin{align}
\begin{array}{cccc}
&\multicolumn{1}{r}{\mathcal{M}_l\mathcal{I}_l\mathcal{O}_l} & {\xrightarrow{\quad u(i_1,t_1,l) }} & \multicolumn{1}{c}{\mathcal{M}_{l+1}\mathcal{I}_{l+1}\mathcal{O}_{l+1}} \\
&&&\\
\multicolumn{2}{l}
{\text{{\scriptsize{$u(i_2,t_2,l+3)$}} } \uparrow \quad } &
 \multicolumn{2}{r}{\quad \quad \quad\downarrow \text{\scriptsize{$u(i_2,t_2,l+1)$}}} \\
 &&&\\
&\multicolumn{1}{l}{\mathcal{M}_{l+3}\mathcal{I}_{l+3}\mathcal{O}_{l+3} }     & \xleftarrow{u(i_1,t_1,l+2)} &\multicolumn{1}{c}{\mathcal{M}_{l+2}\mathcal{I}_{l+2}\mathcal{O}_{l+2}}
\end{array}
\label{eq:vloop}
\end{align} 
stating that the events have to flip at the vertex such that a sequence of up to $3$ single updates reaches each $\mathcal{M}_{l_1}\mathcal{I}_{l_1}\mathcal{O}_{l_1}$ set from any other $\mathcal{M}_{l_2}\mathcal{I}_{l_2}\mathcal{O}_{l_2}$ set associated with that vertex. % is it a recurrence property?

\subsection{AC-LSVMPath algorithm}

We now describe our algorithm. We represent the entire regularization path for the AC-LSVM by the set of vertices $\mathcal{V}$, edges $\mathcal{E}$, and facets $\mathcal{F}$. Let $j,k,l \in \mathbb{N}^+$. Then: 
\begin{equation}
\mathcal{V}=\lbrace v_j : v_j = (C^+,C^-), \langle k \rangle\rbrace
\end{equation}
\begin{equation}
\mathcal{E}=\lbrace e_k : e_k = (i,t), \langle j \rangle, \langle l \rangle\rbrace
\end{equation}
\begin{equation}
\mathcal{F}=\lbrace f_l : f_l = (\mathcal{M}_l \mathcal{I}_l \mathcal{O}_l), \langle k \rangle\rbrace
\end{equation}
where $(\cdot)$ and $\langle \cdot \rangle$ denote attribute and connectivity, respectively, of each element in the $\mathcal{V}\mathcal{E}\mathcal{F}$ sets. 

% structure of the polygon mesh
\paragraph{Ordering} The $\mathcal{V}\mathcal{E}\mathcal{F}$ sets admit the following connectivity structure. Let $\mathcal{V} = \lbrace \mathcal{V}_m \rbrace_{m=1}^M$, $\mathcal{E} = \lbrace \mathcal{E}_m \rbrace_{m=1}^M$, and $\mathcal{F} = \lbrace \mathcal{F}_m \rbrace_{m=1}^M$ be partitioned into $M$ subsets where $v_{j \in j_m} \in \mathcal{V}_m$, $e_{k \in k_m} \in \mathcal{E}_m$, and $f_{l \in l_m} \in \mathcal{F}_m$. The subsets admit a sequential ordering, where $j_m < j_{m+1}$, $k_m < k_{m+1}$, and $l_m < l_{m+1}$, such that edges $e_{k} \in \mathcal{E}_m$ determine the adjacency of facet pairs\footnote{This is also known as \textit{facet-to-facet} property in parametric programming literature (Spjotvold, 2008)} $f_{l_{1,2}} \in \mathcal{F}_m$ or $f_{l_{1}} \in \mathcal{F}_m \wedge f_{l_{2}} \in \mathcal{F}_{m+1}$ while vertices $v_{j} \in \mathcal{V}_m$ determine the intersection of edges $e_{k} \in \mathcal{E}_m$ or $e_k \in \mathcal{E}_m \wedge e_{k} \in \mathcal{E}_{m+1}$. In effect, our algorithm orders facets into a \textit{layer}-like structure.

We define a vertex $v_j$ as an \textit{open vertex} $v_j^o$ when $\vert v_j^o\langle k_m \rangle \vert = 2$ or $\vert v_j^o\langle k_m \rangle \vert = 3$, where $\vert \cdot \vert$ is set cardinality, if the vertex does not lie on neither $C$ axis. We define a \textit{closed vertex} when $\vert v_j^c\langle k_m \rangle \vert = 4$. When $\vert v_j^c\langle k_m \rangle \vert = 3$, the vertex is also closed if it lies on either $C$ axis. Similarly, an edge $e_k$ is called an \textit{open edge} $e_k^{o}$ when $\vert e_k^o\langle j_m \rangle \vert = 1$ and a \textit{closed edge} $e_k^{c}$ when $\vert e_k^c\langle j_m \rangle \vert = 2$. Then, a facet $f_l$ is called an \textit{open facet} when first and last edge in $f_l\langle k_m \rangle$ are unequal; otherwise it is a \textit{closed facet}. Finally, $v_j, e_k, f_l$ are called either \textit{single} $v_j^s, e_k^s, f_l^s$ when they are unique or \textit{replicated} $v_j^r, e_k^r, f_l^r$, otherwise.

We propose to explore the AC-LSVM regularization path in a sequence of layers $m=1,\ldots,M$. We require that $\mathcal{F}_1(\mathcal{M}\mathcal{I}\mathcal{O})$ facet attributes are given at the beginning, where $\vert\mathcal{F}_1 \vert \geq 1 $, $\vert \mathcal{E}_1 \vert \geq 0$, and $\vert \mathcal{V}_1 \vert \geq 0$. An $m$-th layer is then composed of four successive steps.

\textbf{1. Closing open edges and facets} - \texttt{CEF}

For each $\mathcal{M}_l\mathcal{I}_l\mathcal{O}_l$ set, which is attributed to $f_l \in \mathcal{F}_m$, the algorithm separately calls a convex hull routine \texttt{CH}. The routine uses \eqref{eq:H} to compute linear inequality constraints $H_l \geq 0$ creating a convex polytope at $l$. The ordered set of edges $f_l \langle k_m \rangle $, where the first and last edge are open, serve as initial, mandatory constraints in the \texttt{CH} routine. After completion, the routine augments the set $\mathcal{E}_m$ by closed edges $e_{k_m}^c$ and the set $\mathcal{V}_m$ by open vertices $v_{j_m}^o$. 

\textbf{2. Merging closed edges and open vertices} - \texttt{MEV} 

As the \texttt{CH} routine runs for each facet $f_l \in \mathcal{F}_m$ separately, some edges $e_{k_m}^c$ and/or vertices $v_{j_m}^o$ may be replicated, thereby yielding $e_{k_m}^c = e_{k_m}^{rc} \cup e_{k_m}^{sc}$ and $v_{j_m}^o = v_{j_m}^{ro} \cup v_{j_m}^{so}$. 

Notably, a vertex $v_{j_1} \in v_{j_m}^o$ is replicated when another vertex $v_{j_2} \in v_{j_m}^o$ (or other vertices) has the same attribute, i.e. $v_{j_1}(C^+,C^-) = v_{j_2}(C^+,C^-)$. However, we argue that merging vertices into a single vertex based on the distance between them in some metric space may affect the numerical stability of the algorithm. 

On the other hand, a closed edge $e_{k_1} \in e_{k_m}^c$ is replicated by another closed edge $e_{k_2} \in e_{k_m}^c$, when both edges connect a pair of vertices that are both replicated. Replicated edges cannot merge solely by comparing their event attributes $e_k(i,t)$. As they are piecewise continuous in the $(C^+,C^-)$ space, they are not unique. Similarly to vertices though, the edges might be merged by numerically comparing their associated linear constraints, which are only sign-flipped versions of each other, as shown in \eqref{eq:h_alpha0_inv}$-$\eqref{eq:h_L_inv}. However, this again raises concerns about the potential numeric instability of such a merging procedure.

In view of this, we propose a sequential merging procedure that leverages $f_{l \in l_m}(\mathcal{M}_l\mathcal{I}_l\mathcal{O}_l)$ sets, which are both unique in $\mathcal{F}_m$ and discrete. To this end, we first introduce two functions that act on attributes and connectivity of objects $f_l, e_k, v_j$.

Let $I(\mathcal{Q}_1,\mathcal{Q}_2,p)$ be an indexing function that groups $\mathcal{Q}_2$ by assigning labels from set $\mathcal{N}_{q \in \mathcal{Q}} = \left\lbrace q : q \in \mathcal{Q} \right\rbrace$ to $q \in \mathcal{Q}_2$ based on $p(q)$ indexed over $q \in \mathcal{Q}_1 \cup \mathcal{Q}_2$:
\begin{equation}
I(\mathcal{Q}_1,\mathcal{Q}_2,p): \bigcup_{q \in \mathcal{Q}_1 \cup \mathcal{Q}_2} p(q) \rightarrow \mathcal{N}_{q \in \mathcal{Q}_1}^{\vert \mathcal{Q}_2 \vert \times 1}
\end{equation}

Let $R(\mathcal{Q}_1,\mathcal{Q}_2,p)$ then be a relabeling function that assigns labels from $\mathcal{N}(q_2)$ to $p(q_1)$ indexed over $q_1 \in \mathcal{Q}_1$:
\begin{equation}
R(\mathcal{Q}_1,\mathcal{Q}_2,p): \forall_{q_1 \in \mathcal{Q}_1} \forall_{q_2 \in \mathcal{Q}_2} ~~\mathcal{N}(q_2) \rightarrow p(q_1)
\end{equation}

The algorithm commences the merging procedure by populating initially empty set $\mathcal{F}_{m+1}$  with facets $f_{l \in l_{m+1}}$ that are obtained by separately updating \eqref{eq:update} the facets $f_{l \in l_m}$ through the events attributed to each edge $e_k \in e_{k_m}^c$. Note, however, that replicated edges $e_{k_{m}}^{rc}$ will produce facet attributes in $\mathcal{F}_{m+1}$ that replicate facet attributes from the preceding layer. Moreover, single edges $e_{k_{m}}^{sc}$ may as well produce replicated facet attributes in the current layer. Hence, we have that $f_{l \in l_{m+1}}^r(\mathcal{M}_l\mathcal{I}_l\mathcal{O}_l) \subseteq f_{l \in l_{m}}^s(\mathcal{M}_l\mathcal{I}_l\mathcal{O}_l) \cup f_{l \in l_{m+1}}^s(\mathcal{M}_l\mathcal{I}_l\mathcal{O}_l)$. 

In order to group facets into single and replicated $\mathcal{M}\mathcal{I}\mathcal{O}$ sets, the algorithm indexes facet attributes with $I(l_m \cup l_{m+1}^s, l_{m+1}^s \cup l_{m+1}^r, f_l(\mathcal{M}_l\mathcal{I}_l\mathcal{O}_l))$ based on their equality. Then, relabeling facet-edge connectivities of edges $e_{k_m}^c$ with $R(k_m^c, l_{m+1}^s \cup l_{m+1}^r, e_k\langle l \rangle)$ allows for indexing the connectivities with $I(k_m^{sc},k_m^{sc} \cup k_m^{rc},e_k\langle l \rangle)$ also based on their equality. Having indicated single and replicated edges, the algorithm relabels edge-vertex connectivities of vertices $v_{j_m}^o$ with $R(j_m^o,k_{m}^c, v_j\langle k \rangle)$. 

Note that there are two general vertex replication schemes. Vertices, which indicate the occurrence of joint events, can be replicated when two facets connect (i) through an edge (i.e., vertices share replicated edge) or (ii) through a vertex (i.e., vertices connect only to their respective, single edges). At this point of the merging procedure, a vertex $v_{j_m}^o$ is associated indirectly through edges $e_{k_m}^{sc}$ with two facets $f_l \in f_{l \in l_{m}}^s \cup f_{l \in l_{m+1}}^s$, when it lies on either $C$ axis, or three facets $f_l \in f_{l \in l_{m}}^s \cup f_{l_{1,2} \in l_{m+1}}^s$, otherwise. 

Two vertices lying on, say, $C^+$ axis are replicated when their respective edges share a facet and are attributed the events that refer to the same, negative point $\lbrace x_i, y_i=-1\rbrace$, but yet that have opposite event types, i.e. $t_1 = \neg~t_2$. This condition follows directly from \eqref{eq:h_alpha0}$-$\eqref{eq:h_L}, as $\alpha_i=C^-=0$ at $(C^+,0)$. Conversely, when vertices lie on $C^-$ axis, they are replicated when their edges have events referring to the positive point, $\lbrace x_i, y_i=+1\rbrace$. Then, two vertices lying on neither $C$ axis are replicated when their respective edges are associated with two common facets and equal joint events. Hence, vertices are indexed with $I(j_m^{so},j_m^{so} \cup j_m^{ro}, v_j\langle k \rangle\rightarrow e_k \langle l \rangle \cup e_k(i,t))$ based on the equality of edge-facet connectivities along with edge attributes. Alternatively, should the joint events be unique, the vertices could then be merged solely by comparing these events. Showing that joint events being unique is true or false, i.e. two $1$D event paths can intersect only at a single point in the entire $(C^+,C^-)$ space, is an interesting future work.

Having grouped facets $f_{l \in l_{m+1}}$, edges $e_{k \in k_{m}}^c$, and vertices $v_{j \in j_{m}}^o$, now the algorithm can merge facet-edge connectivities $f_l \langle k \rangle$ of the replicated facets, prune replicated edges $e_k^r$, and merge vertex-edge connectivites $v_j \langle k \rangle$ of the replicated vertices. 

Being left with only single facets $f_{l \in l_{m+1}}^s$, edges $e_{k \in k_{m}}^{sc}$, and vertices $v_{j \in j_{m}}^{so}$, the algorithm relabels with $R(k_m^{so} \cup k_m^{sc}, j_m^o, e_k\langle j \rangle)$ the edge-vertex connectivities of single, open edges $e_{k_m}^{so}$ and single, closed edges $e_{k_m}^{sc}$ intersecting with $e_{k_m}^{so}$. Finally, the algorithm relabels with $R(l_m \cup l_{m+1}, k_m^c, f_l\langle k \rangle)$ the facet-edge connectivites of facets from the preceding and current layer.

\textbf{3. Closing open vertices} - \texttt{CV} 

In this step, the algorithm closes the vertices $v_{j_m}^{so}$ by attaching open edges. Specifically, by exploiting the piecewise continuity of events at vertices, the algorithm populates the $\mathcal{E}_{m+1}$ set with open edges $e_{k_{m+1}}^o$, such that a vertex $v_{j_m}^{sc}$ now connects either to (i) $3$ edges, when it lies on, say, $C^+$ axis and connects to event edge with associated positive point, or to (ii) $4$ edges when it lies on neither axis. 

Using the vertex loop property \eqref{eq:vloop}, the algorithm then augments the set $\mathcal{F}_{m+1} = \mathcal{F}_{m+1} \cup f_{l_{m+1}}^a$ with additional facets $f_{l_{m+1}}^a$ such that now the closed vertex $v_{j_m}^{sc}$ connects indirectly through its edges $e_k \in e_{k_m}^o \cup e_{k_m}^c \cup e_{k_{m+1}}^o$ to facets $f_l \in f_{l_m} \cup f_{l_{m+1}}$ and additionally up to $1$ new facet $f_l \in f_{l_{m+1}}^a$. 

There are several advantages for generating open edges. Firstly, augmenting the initialization list of edges during the call to the \texttt{CH} routine reduces the number of points for processing, with computational load $\mathcal{O}(N-\vert f_l\langle k_m \rangle\vert)$. Secondly, each vertex generates up to two single open edges. However, there can be two single vertices that generate the same open edge thereby merging the $1$D path of an event. In this case, both open edges are merged into a single closed edge and the facet is closed without processing it with \texttt{CH} routine. This merging step is described next.

\textbf{4. Merging open edges and facets} - \texttt{MEF}

As open edges and their facets, which are generated in step $3$, can also be single or replicated, step $4$ proceeds similarly to step $2$. 

The algorithm indexes additional facets with $I(l_{m+1}^s \cup l_{m+1}^{sa}, l_{m+1}^{sa} \cup l_{m+1}^{ra}, f_l(\mathcal{M}_l\mathcal{I}_l\mathcal{O}_l))$ and relabels the open edge connectivities with $R(k_{m+1}^o, l_{m+1}^{sa} \cup l_{m+1}^{ra}, e_k\langle l \rangle)$. Then, the algorithm indexes these connectivites with $I(k_{m+1}^{so},k_{m+1}^{so} \cup k_{m+1}^{ro},e_k\langle l \rangle)$ and merges edge-vertex $e_k \langle j \rangle$ and facet-edge $f_l \langle k \rangle$ connectivities. Finally, the algorithm relabels with $R(l_{m+1} \cup l_{m+1}^a, k_{m+1}^o, f_l\langle k \rangle)$ the facet-edge connectivity of all facets in $\mathcal{F}_{m+1}$ and returns to step $1$.

\paragraph{Termination} The algorithm terminates at $M$-th layer, in which the \texttt{CH} routine for $\mathcal{M}\mathcal{I}\mathcal{O}$ sets of all facets in $\mathcal{F}_M$, where $\vert\mathcal{F}_M\vert \geq1$, produces \textit{open polytopes} in the $(C^+,C^-)$ space. 
% BUT: Is here the last merging of edges and verts required?

\paragraph{Special cases} As mentioned in (Hastie et al., 2004), two special case events may occur after closing facet $f_l$ and set updating \eqref{eq:update}. When (i) replicated data points $\left\lbrace \mathbf{x}_i,y_i\right\rbrace $ exist in the dataset and enter the margin, or (ii) single points simultaneously project onto the margin such that $\vert \mathcal{M}_{l+1}\vert > d$, then the matrix $\mathbf{X}_{\mathcal{M}_{l+1}}^T\mathbf{X}_{\mathcal{M}_{l+1}}$ becomes singular and thus not invertible, yielding non-unique paths for some $\alpha_i$. In contrast to (Hastie et al., 2004), note that the case (ii) is likely to occur in the considered LSVM formulation (1)$-$\eqref{eq:bias_augm} as the positive and negative data points span up to $d-1$ subspace after being affine transformed, yielding e.g. parameters $\boldsymbol{\beta}= \left[ \mathbf{0} ,1/B \right]$.
%\left[ \widehat{\boldsymbol{\beta}}, b\right] = 

In the context of our algorithm, both cases (i)$-$(ii) are detected at $f_l$ when the matrix $\widehat{\mathbf{H}}_{n \times 3}$ formed of $n \geq 3$ constraints \eqref{eq:h_alpha0}$-$\eqref{eq:h_L} associated with these points either has $rank(\widehat{\mathbf{H}}) = 1$, producing multiple events at an edge denoted by constraints that are identical up to positive scale factor, or has $rank(\widehat{\mathbf{H}}) = 2$, producing multiple joint events at a vertex denoted by constraints that intersect at the same point. 

We propose the following procedure for handling both special cases. Namely, when some facets $f_l \in f_{l_m}$ close with edges having multiple events or with vertices having multiple joint events that would lead to cases (i)$-$(ii), the algorithm moves to step $2$, as it can obtain facet updates in these special cases. However, it skips step $3$ for these particular facets. While we empirically observed that such vertices close with $2$ edges having multiple joint events, it is an open issue how to generate open edges in this case. Instead, during successive layers, step $2$ augments the list of facets, edges, and vertices by the ones associated to (i)$-$(ii) for indexing and relabeling them with respect to successive ones that will become replicated in further layers. In effect, our algorithm 'goes around' these special case facets and attempts to close them by computing adjacent facets. However, the path for $\boldsymbol{\alpha}$ in these cases is not unique and remains unexplored. Nevertheless, our experiments suggest that unexplored regions occupy relatively negligibly small area in the hyperparameter space.

When the algorithm starts with all points in $\mathcal{I}$ and either case (i)$-$(ii) occurs at the initial layers, the exploration of the path may halt\footnote{The path exploration will halt when these cases occur as the data points that are the first to enter the margin; and more generally, when $1$D multiple event paths referring to these cases will go to both $C$ axis, instead of to one $C$ axis and to infinity.} due to the piecewise continuity of the (multiple) events. A workaround can then be to run a regular LSVM solver at yet unexplored point $(C^+,C^-)$, obtain $\mathcal{M}\mathcal{I}\mathcal{O}$ sets, and extract convex polytope to restart the algorithm.

Our future work will focus on improving our tactics for special cases. We posit that one worthy challenge in this regard is to efficiently build the entire regularization path in $N$-dimensional hyperparameter space.

\paragraph{Computational complexity} 

Let $\vert \widehat{\mathcal{M}} \vert$ be the average size of a margin set for all $l$, let $\vert \widehat{\mathcal{F}} \vert$ be the average size of $\mathcal{F}_m$. Then, the complexity of our algorithm is $\mathcal{O}(M \vert \widehat{\mathcal{F}} \vert (N + \vert \widehat{\mathcal{M}} \vert^3))$, where $\vert \widehat{\mathcal{M}} \vert^3$ is the number of computations for solving \eqref{eq:h_alpha0} (without inverse updating/downdating (Hastie et al., 2004)) and we hid constant factor $2$ related to convex hull computation. However, note that typically we have $\vert \widehat{\mathcal{M}} \vert \ll N$. In addition, we \textit{empirically} observed that $M \approx N$ (but cf. (Gartner et al., 2012)), so that the number of layers $m$ approximates dataset size. Our algorithm is sequential in $M$ but parallel in $\vert \widehat{\mathcal{F}} \vert$. Therefore, the complexity of a parallel implementation of the algorithm can drop to $\mathcal{O}(N^2 + N\vert \widehat{\mathcal{M}} \vert^3)$. Finally, at each facet, it is necessary to evaluate \eqref{eq:h_alpha0}. But then the evaluation of constraints \eqref{eq:h_alphaC1}$-$\eqref{eq:h_R} can be computed in parallel, as well. While this would lead to further reduce the computational burden, memory transfer remains the main bottleneck on modern computer architectures. 

Our algorithm partitions the sets $\mathcal{F}$, $\mathcal{E}$, $\mathcal{V}$ into a \textit{layer}-like structure such that our two-step merging procedure requires access to objects only from layer pairs $m$ and $m+1$ and not to preceding layers\footnote{When the algorithm encounters special cases at $m$, it requires access to $f_l$, $e_k$, $v_j$ objects related to these cases even after $m_1 > m+1$ layers, but the number of these objects is typically small.}. In effect, the algorithm only requires $\mathcal{O}(\vert \widehat{\mathcal{F}} \vert + \vert \widehat{\mathcal{E}} \vert + \vert \widehat{\mathcal{V}} \vert)$ memory to cache the sets at $m$, where $\vert \widehat{\mathcal{E}} \vert$ and $\vert \widehat{\mathcal{V}} \vert$ are average edge and vertex subset sizes of $\mathcal{E}_m$ and $\mathcal{V}_m$, respectively.

\begin{figure*}[h]
\vspace{.3in}
\centerline{\fbox{
\includegraphics[width=\linewidth, height=8.0cm] {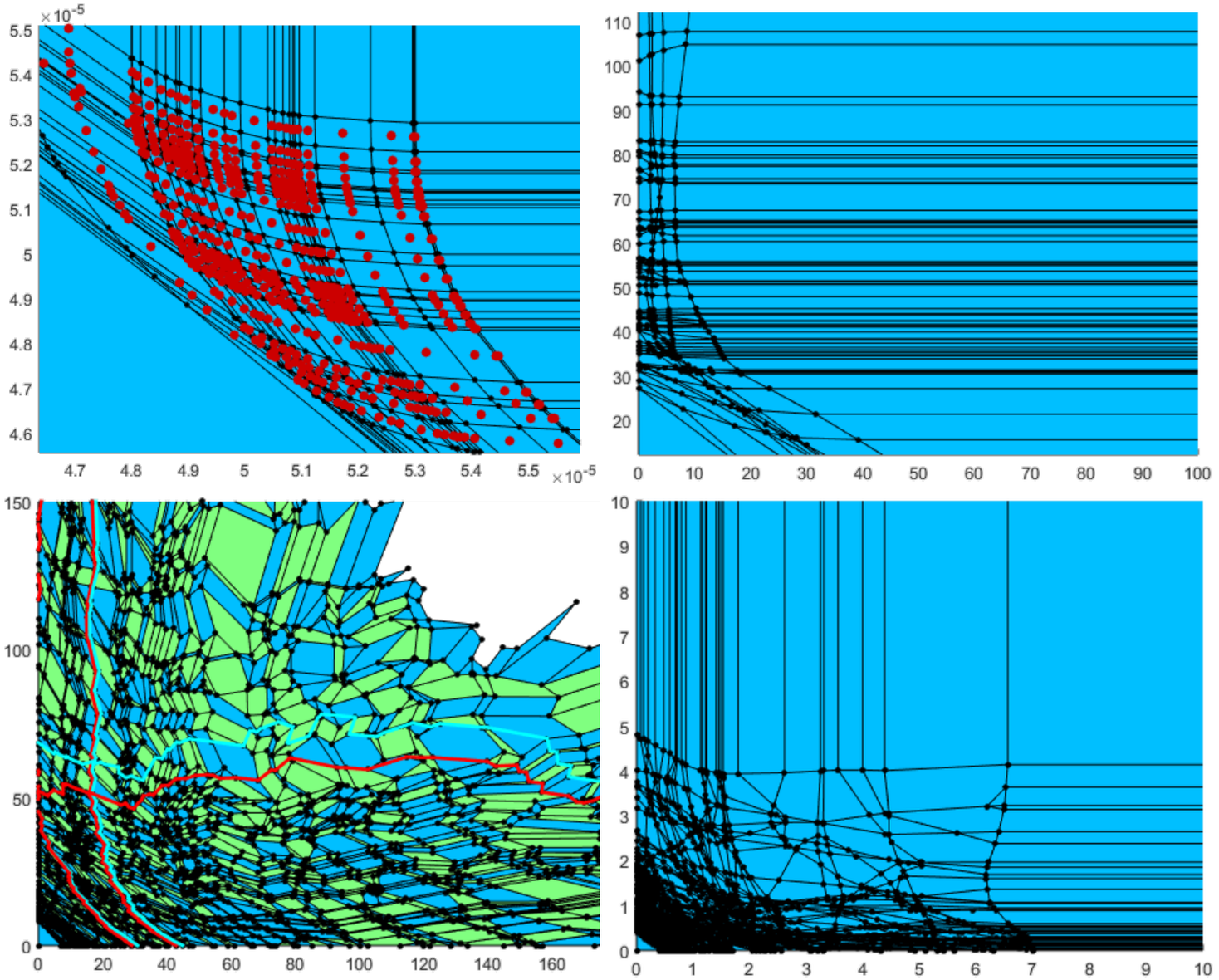} 
}}
\vspace{.3in}
\caption{Visualization of the entire regularization path for the AC-LSVM. Experiments (i)-(iv) are shown in counterclockwise order. In (i) we show a portion of the entire regularization path, where red dots indicate facet means. In (ii) we show intertwined layers of facets up to some layer $m$ (blue and green) and 1D event paths of several points (cyan - event $t=0$ and red - event $t=1$). In (iii) and (iv) we show the entire regularization path.}
\end{figure*}

\begin{figure*}[h]
\vspace{.3in}
\centerline{\fbox{
\includegraphics[width=\linewidth, height=8.0cm]{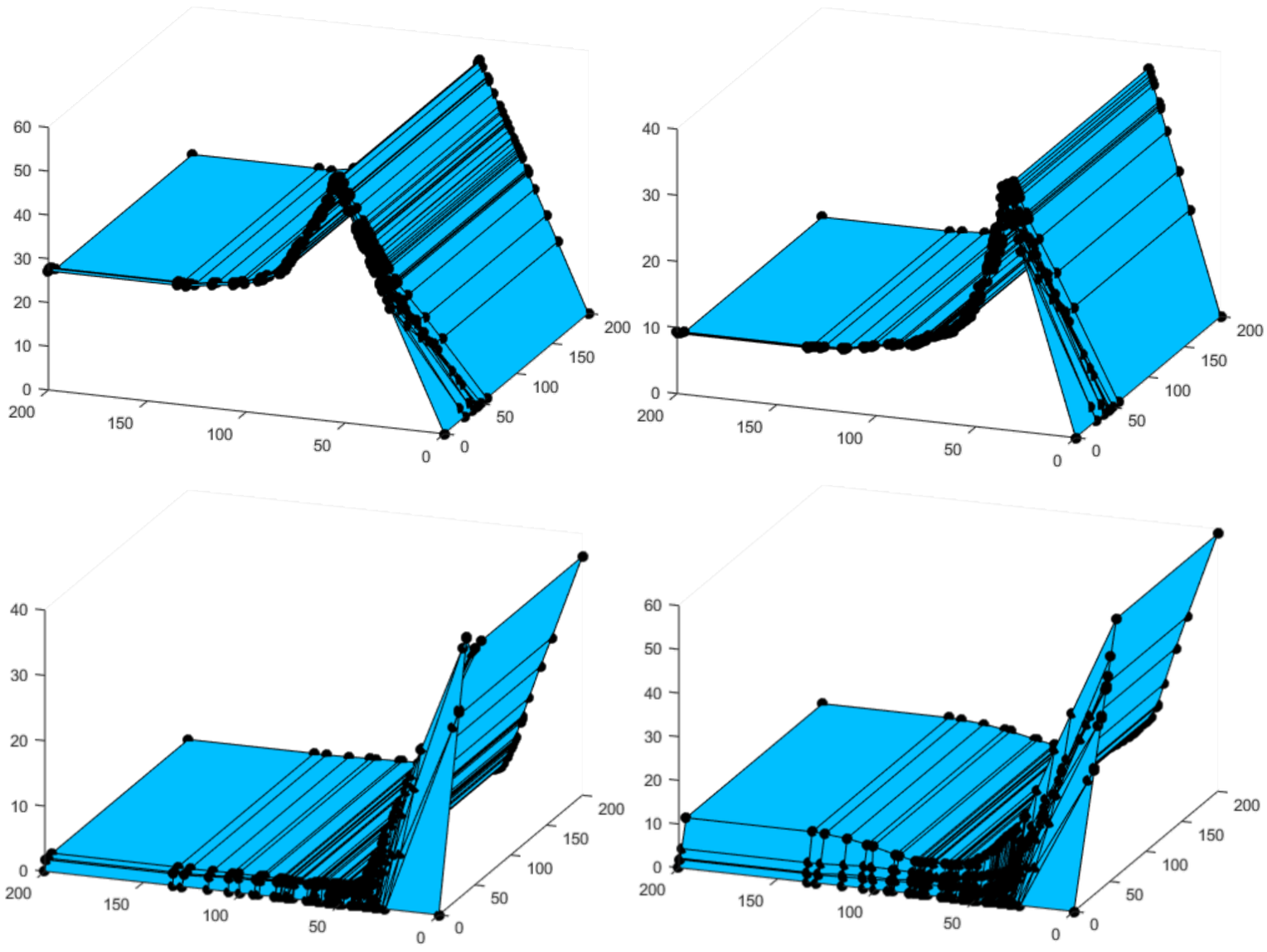} 
}}
\vspace{.3in}
\caption{Visualization of the entire regularization paths for several Langrange multipliers $\alpha_i$ for experiment (iv).}
\end{figure*}

\section{Numerical experiments}

In this section, we evaluate our AC-LSVMPath algorithm described in section \ref{sec:algo}. We conduct three numerical experiments for exploring the two-dimensional path of assymetric-cost LSVMs on synthetic data. We generate samples from a gaussian distribution $\mathcal{N}(\mu, \sigma^2)$ for (i) a small dataset with large number of features $N \ll d$, (ii) a large dataset with small number of features $N \gg d$, and (iii) a moderate size dataset with moderate number of features $N = d$. 

We also build two-dimensional regularization path when input features are sparse (iv). We use off-the-shelf algorithm for training flexible part mixtures model (Yang \& Ramanan, 2013), that uses positive examples from Parse dataset and negative examples from INRIA's Person dataset (Dalal \& Triggs, 2006). The model is iteratively trained with hundreds of positive examples and millions of hard-mined negative examples. We keep original settings. The hyperparameters are set to $C^+=0.004$ and $C^-=0.002$ to compensate for imbalanced training (Akbani et al., 2004). 

For experiments (i)$-$(iv), we have the following settings: (i)  $d=10^6$, $N^+=25$, $N=50$, (ii) $d=2$, $N^+=50$, $N=100$, (iii) $d=10^2$, $N^+=50$, $N=100$, (iv) $d\approx 10^5$, $N^+=10$, $N=50$. We set $B=0.01$ in all experiments, as in (Yang \& Ramanan, 2013). The results are shown in Fig. 1 and Fig. 2.

\section{Conclusions}

This work proposed an algorithm that explores the entire regularization path of asymmetric-cost linear support vector machines. The events of data concurrently projecting onto the margin are usually considered as special cases when building one-dimensional regularization paths while they happen repeatedly in the two-dimensional setting. To this end, we introduced the notion of joint events and illustrated the set update scheme with vertex loop property to efficiently exploit their occurrence during our iterative path exploration. Finally, as we structure the path into successive layers of sets, our algorithm has modest memory requirements and can be locally parallelized at each layer of the regularization path. Finally, we posit that extending our algorithm to the entire $N$-dimensional regularization path would facilitate processing of further special cases.

\subsubsection*{References}

Japkowicz, N., Stephen, S. (2002). The class imbalance problem: A systematic study. {\it Intelligent Data Analysis}, 6(5), 429-449

Bergstra, J., Bengio, Y. (2012). Random search for hyper-parameter optimization. {\it The Journal of Machine Learning Research}, 13(1), 281-305

Bach, F. R., Heckerman, D., Horvitz, E. (2006). Considering cost asymmetry in learning classifiers. {\it The Journal of Machine Learning Research}, 1713-1741

Hastie, T., Rosset, S., Tibshirani, R., Zhu, J. (2004). The entire 
regularization path for the support vector machine. {\it The Journal of 
Machine Learning Research}, 1391-1415

Hsieh, C. J., Chang, K. W., Lin, C. J., Keerthi, S. S., Sundararajan, S. (2008). A dual coordinate descent method for large-scale linear SVM. In { \it Proceedings of the 25th International Conference on Machine learning}, 408-415

Spjotvold, J. (2008). Parametric programming in control
theory. {\it PhD Thesis}

Fan, R. E., Chang, K. W., Hsieh, C. J., Wang, X. R., Lin, C. J. (2008). LIBLINEAR: A library for large linear classification. {\it The Journal of Machine Learning Research}, 9, 1871-1874

Shalev-Shwartz, S., Singer, Y., Srebro, N., Cotter, A. (2011). Pegasos: Primal estimated sub-gradient solver for SVM. {\it Mathematical programming}, 127(1), 3-30

Joachims, T. (2006). Training linear SVMs in linear time. In {\it ACM SIGKDD International Conference on Knowledge Discovery and Data mining}, 217-226

Gartner, B., Jaggi, M., Maria, C. (2012). An exponential lower bound on the complexity of regularization paths. {\it Journal of Computational Geometry}, 3(1), 168-195

Karasuyama, M., Harada, N., Sugiyama, M., Takeuchi, I. (2012). Multi-parametric solution-path algorithm for instance-weighted support vector machines. {\it Machine learning}, 88(3), 297-330

Pedregosa, F., Varoquaux, G., Gramfort, A., Michel, V., Thirion, B., Grisel, O., et al., Duchesnay, E. (2011). Scikit-learn: Machine learning in Python. {\it The Journal of Machine Learning Research}, 12, 2825-2830

Klatzer, T., Pock, T. (2015). Continuous Hyper-parameter Learning for Support Vector Machines. In {\it Computer Vision Winter Workshop}

Chu, B. Y., Ho, C. H., Tsai, C. H., Lin, C. Y., Lin, C. J. (2015). Warm Start for Parameter Selection of Linear Classifiers. {\it  ACM SIGKDD International Conference on Knowledge Discovery and Data Mining}, 149-158

Snoek, J., Larochelle, H., Adams, R. P. (2012). Practical Bayesian optimization of machine learning algorithms. In {\it Advances in Neural Information Processing Systems}, 2951-2959

Ong, C. J., Shao, S., Yang, J. (2010). An improved algorithm for the solution of the regularization path of support vector machine. {\it IEEE Transactions on Neural Networks}, 21(3), 451-462

Akbani, R., Kwek, S., Japkowicz, N. (2004). Applying support vector machines to imbalanced datasets. In {\it Machine Learning: ECML}, 39-50

Dalal, N., Triggs, B. (2005). Histograms of oriented gradients for human detection. In {\it Computer Vision and Pattern Recognition},886-893

Boser, B. E., Guyon, I. M., Vapnik, V. N. (1992). A training algorithm for optimal margin classifiers. In {\it Proceedings of the fifth annual workshop on Computational learning theory}, 144-152

Avis, D., Bremmer, D., Seidel, R. (1997). How good are convex hull algorithms?. {\it Computational Geometry}, 7(5), 265-301

\end{document}